\documentclass[sigconf]{acmart}
\AtBeginDocument{%
  }

\usepackage{subcaption}
\usepackage{multirow}
\usepackage{soul}
\usepackage{algorithm}
\usepackage{algpseudocode}

\renewcommand\footnotetextcopyrightpermission[1]{}

\begin{document}

\title{CSI-JEPA: Towards Foundation Representations for Ubiquitous Wi-Fi Sensing with Minimal Supervision}

\author{Xuanhao Luo}
\email{xluo26@ncsu.edu}
\affiliation{%
  \institution{North Carolina State University}
  \state{Raleigh, NC}
  \country{USA}
}

\author{Zhizhen Li}
\email{zli92@ncsu.edu}
\affiliation{%
  \institution{North Carolina State University}
  \state{Raleigh, NC}
  \country{USA}
}

\author{Yuchen Liu}
\email{yuchen.liu@ncsu.edu}
\affiliation{%
  \institution{North Carolina State University}
  \state{Raleigh, NC}
  \country{USA}
}

\begin{abstract}
Channel state information (CSI) provides a widely available sensing modality for human and environment perception, but existing CSI sensing models usually rely on task-specific supervised training and require substantial labeled data for each task, device, user, or environment. This limits their scalability in practical deployments where unlabeled CSI is abundant but labeled data is costly to collect. In this paper, we present CSI-JEPA, a self-supervised predictive representation learning framework for label-efficient and transferable Wi-Fi sensing. CSI-JEPA learns reusable temporal-spectral representations from unlabeled CSI samples by predicting latent features of masked channel regions from visible context. To better match the physical structure of CSI, it jointly tokenizes CSI along the time and subcarrier dimensions and introduces a channel variation-aware masking strategy that prioritizes regions with stronger local channel dynamics. After pretraining, the encoder is frozen and adapted to downstream sensing tasks using lightweight task-specific adapters. We evaluate CSI-JEPA on seven real-world Wi-Fi sensing tasks spanning diverse objectives and deployment settings. The results show that CSI-JEPA improves downstream sensing performance over competitive baselines, achieving up to 10.64 percentage points (pp) mean accuracy gain over the supervised Transformer and matched-budget label savings of up to 95.0\%.
Across unseen tasks, environments, hardware platforms, and CSI organizations, frozen CSI-JEPA improves downstream adaptation by up to 22.92 pp over non-pretrained baselines, demonstrating strong representation transferability.
\end{abstract}


\maketitle

\pagestyle{plain}

\section{Introduction}
Wi-Fi sensing has emerged as a promising paradigm for ubiquitous and device-free perception \cite{li2021deep, chen2023cross, ma2019wifi}. 
By analyzing wireless channel perturbations caused by human motion, breathing, body presence, and environmental changes, Wi-Fi systems can support a wide range of sensing applications without requiring targets to carry dedicated sensors \cite{wu2022wi, liu2019wireless}. 
Channel state information (CSI), which captures fine-grained channel responses across OFDM subcarriers and packet time indices, is particularly attractive for wireless sensing because it can be readily obtained from commodity Wi-Fi devices, especially with the growing support from the emerging IEEE 802.11bf protocol~\cite{ropitault2024ieee, du2024overview}.

\begin{figure}[h]
    % \vspace{-1em}
    \centerline{\includegraphics[scale=1.15]{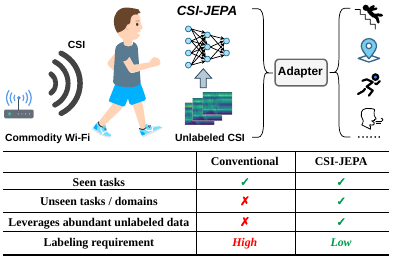}}
        % \vspace{-0.6em}
    \caption{ CSI-JEPA learns reusable representations from unlabeled CSI and supports label-efficient adaptation to seen and unseen sensing domains.}
    \label{1}
    % \vspace{-1em}
\end{figure}

Despite this promise, practical Wi-Fi sensing systems still face a major scalability challenge. As illustrated in Fig. \ref{1}, most existing CSI-based sensing frameworks are trained in a supervised and task-specific manner \cite{li2026beamforming, liu2015contactless, cheng2019walls, li2025bfmloc, sapiezynski2017inferring}, requiring substantial labeled CSI data for each sensing task, user group, device configuration, or environment.
However, labeled CSI, i.e., CSI measurements paired with task-specific ground-truth annotations, is often costly to collect because it requires synchronized sensing activities and controlled experimental procedures~\cite{yang2022autofi}. Besides, the existing labeled data are typically collected using heterogeneous hardware platforms and measurements, limiting cross-dataset interoperability and reusability~\cite{ram2026aren, gong2026towards}.
In contrast, unlabeled CSI can be continuously and readily collected during normal Wi-Fi operation. 
This mismatch between abundant unlabeled CSI and scarce labeled data motivates learning a reusable CSI representation that can be adapted to downstream sensing tasks with minimal supervision.
More importantly, a reusable representation should remain effective beyond its pretraining domain, including unseen sensing tasks, environments, and hardware platforms.

Self-supervised representation learning provides a natural way to address this problem \cite{radwan2025tutorial}. 
However, many existing self-supervised methods rely on consistency learning \cite{yang2022autofi} or masked reconstruction \cite{zhu2026fm, jiang2026csi}, which may emphasize input-level similarity or low-level CSI recovery rather than task-relevant channel dynamics.
Recent advances in the joint-embedding predictive architecture (JEPA) learn representations by predicting target embeddings from context embeddings in latent space, rather than reconstructing raw inputs~\cite{lecun2022path}. 
By shifting the pretext task from raw signal reconstruction to latent-space prediction, JEPA provides a promising way to learn reusable representations that are more aligned with downstream predictive objectives.
However, directly applying JEPA to CSI-based sensing is non-trivial. 
CSI measurements form a structured temporal-subcarrier field, where the temporal axis reflects motion evolution and physiological dynamics, while the subcarrier axis reflects frequency-selective fading and multipath correlations. 
Therefore, the mask-and-predict strategy in JEPA should preserve and exploit this physical structure.
A generic masking strategy may ignore informative channel variations, whereas overly aggressive time-only or subcarrier-only masking may remove too much information needed for predictive learning.

In this paper, we present CSI-JEPA, a self-supervised predictive representation learning framework tailored for label-efficient and transferable Wi-Fi sensing. 
CSI-JEPA tokenizes CSI amplitude windows into temporal-subcarrier patch tokens and learns reusable representations through masked latent prediction. 
To better match the physical structure of CSI, we introduce a channel variation-aware masking strategy that estimates local channel dynamics from temporal and subcarrier-domain variations and samples predictive target regions with stronger channel dynamics. 
Notably, the model is pretrained solely on unlabeled CSI samples without any supervision. 
After pretraining, the encoder is frozen and transferred to multiple Wi-Fi sensing tasks using lightweight task-specific adapters. 
This design separates representation learning from task adaptation and allows the same pretrained backbone to support diverse sensing objectives with limited labeled supervision, including adaptation to unseen tasks and deployment domains.
In particular, CSI-JEPA provides a reusable predictive representation layer between PHY-layer CSI acquisition and application-layer sensing inference. 
By operating on CSI measurements available in existing Wi-Fi systems, CSI-JEPA can serve as a protocol-compatible sensing primitive for future integrated WLAN communication and sensing services as envisioned by IEEE 802.11bf protocol~\cite{meneghello2023toward}. 
Its input modality is compatible with channel measurements obtained through standard Wi-Fi sounding and sensing procedures \cite{yi2024bfmsense, sahoo2024sensing}, allowing routinely collected unlabeled CSI measurements to be reused for self-supervised representation learning. 
The main contributions of this paper are summarized as follows.
\begin{itemize}
    \item \textbf{Learning More from Unlabeled Data}: We propose CSI-JEPA, the first joint-embedding predictive representation learning framework for label-efficient and transferable Wi-Fi sensing. CSI-JEPA learns reusable temporal-subcarrier representations from unlabeled CSI samples and transfers the frozen encoder to downstream sensing tasks with lightweight model adaptation.

    \item \textbf{Masking What Matters Most}: We introduce a channel variation-aware masking strategy tailored to unlabeled CSI data. Instead of uniformly sampling masked local channel segments, the proposed strategy estimates local temporal and subcarrier-domain channel variations and adaptively selects predictive target regions with potentially stronger dynamics.

    \item \textbf{Better Performance, Fewer Labels}: We conduct a comprehensive evaluation on seven real-world Wi-Fi sensing tasks under different labeled budgets. CSI-JEPA consistently improves downstream sensing performance over raw-feature baselines, supervised Transformer training, and reconstruction-based self-supervised pretraining. Compared with the strongest Transformer model, CSI-JEPA achieves up to 10.64 percentage points (pp) mean accuracy gain and up to 14.38 pp mean F1 gain, while reducing the matched labeled budget by up to 95.0\%.

    \item \textbf{Generalizing to the Unseen}: We investigate out-of-distribution representation transfer to unseen sensing tasks, environments, hardware, and CSI organization. Without updating the pretrained encoder, CSI-JEPA improves downstream adaptation by up to 22.92 pp over the corresponding non-pretrained baseline, demonstrating that the learned representation remains reusable beyond its pretraining domain.

\end{itemize}

\section{Related Works}

\textbf{Supervised and multi-task wireless sensing.}
Wireless channel measurements have been widely used in networking and communication systems, supporting tasks such as channel estimation~\cite{feng2024deep}, beam management~\cite{li2025contextual}, and radio map estimation~\cite{luo2025denoising}.
In Wi-Fi sensing, channel-related measurements have further enabled device-free perception tasks such as localization \cite{li2026beamforming}, respiration monitoring \cite{liu2015contactless}, user identification \cite{cheng2019walls}, and proximity estimation \cite{sapiezynski2017inferring}. 
Most existing CSI sensing systems rely on task-specific feature extraction or supervised deep learning models trained for a particular sensing objective, environment, device setup, or user group. 
Recent work has also started to explore unified multi-task sensing models. For instance,
LLM4WM adapts a pretrained language model to multiple channel-associated communication tasks through multi-task adapters and MoE-LoRA \cite{liu2025llm4wm}.
MMSense adapts a vision-based foundation model for multi-modal and multi-task wireless sensing by integrating image, radar, LiDAR, and textual inputs for channel, human, and environment sensing tasks~\cite{li2026mmsense}. 
While these methods can achieve strong performance when sufficient labeled data are available, they often require costly data collection, device- and environment-specific calibration, and repeated retraining to adapt to each new deployment.

\textbf{Self-supervised learning models for CSI.}
Self-supervised learning (SSL) has recently been explored as a way to reduce labeling requirements in wireless sensing and communication systems \cite{radwan2025tutorial}. 
AutoFi learns transferable CSI representations from randomly collected unlabeled CSI samples using consistency-based SSL with mutual-information and geometric structural objectives, enabling few-shot human gesture and gait recognition~\cite{yang2022autofi}.
AM-FM pretrains a Wi-Fi sensing foundation model on large-scale unlabeled CSI using a hybrid self-supervised objective that combines contrastive learning, masked reconstruction, and physics-informed autocorrelation prediction~\cite{zhu2026fm}. 
CSI-MAE applies masked autoencoder pretraining to complex CSI generated from 3GPP channel models, learning channel representations for channel extrapolation, channel feedback, and user positioning~\cite{jiang2026csi}. 
These works demonstrate the potential of self-supervised pretraining for learning transferable wireless representations. 
{However, reconstruction-based objectives train the model to recover raw CSI values, which can make the learned representation sensitive to low-level amplitude variations or channel-specific details that may not always be the most discriminative factors for downstream sensing tasks.} 

\textbf{JEPA for wireless networks.}
Different from traditional supervised learning methods that rely on task-specific labels and existing self-supervised CSI processing that often use view consistency or raw reconstruction, JEPA learns representations by predicting target embeddings from context embeddings in latent space rather than reconstructing raw inputs~\cite{lecun2022path}.
It has since been extended to practical self-supervised representation learning across images~\cite{assran2023self}, videos~\cite{bardes2023v}, and vision-language signals~\cite{chen2025vl}. 
JEPA-style predictive representation learning has also started to appear in wireless network systems. 
WirelessJEPA~\cite{chu2026wirelessjepa} learns general-purpose representations from raw multi-antenna I/Q streams using masked latent prediction over antenna-time grids, and evaluates the learned encoder on RF-centric tasks such as modulation classification, AoA estimation, and RF fingerprinting. 
While raw I/Q signals provide fine-grained physical-layer information, collecting synchronized multi-antenna I/Q streams often requires dedicated radio hardware and controlled measurement setups. 
Recent work has also applied JEPA to CSI trajectory modeling by learning velocity-conditioned latent channel dynamics, where future channel-chart embeddings are predicted from current CSI representations and user velocity~\cite{chaaya2024learning}. 
Follow-up work further structures these latent transitions using homomorphic world models with Lie algebra-based action operators to improve geometric consistency and compositional rollout prediction~\cite{naoumi2026structured}. 
Beyond CSI and I/Q signals, JEPA-MSAC~\cite{zheng2026jepa} applies temporal block-masked JEPA to multimodal sensing-assisted communications, using vision, radar, LiDAR, GPS, and RF beam-level power measurements to support localization, beam prediction, and RSSI prediction. 
Different from these prior efforts, our CSI-JEPA focuses on label-efficient Wi-Fi sensing from temporal-subcarrier CSI amplitude windows and introduces channel variation-aware target selection for JEPA pretraining without requiring hard-to-obtain I/Q streams, specialized multimodal sensing hardware, explicit velocity measurements, trajectory-level supervision, or any position annotations.

\section{System Model and Problem Formulation}

\subsection{CSI-based Sensing System}

We consider a general Wi-Fi sensing system consisting of wireless transmitters and receivers deployed in an indoor environment.
A transmitter sends Wi-Fi packets to a receiver at regular or application-dependent intervals, and the receiver estimates CSI from received packets.
CSI characterizes the complex channel response between the transmitter and receiver over multiple subcarriers and packet time indices.
For systems with multiple antennas, receiver streams, or channel views, CSI can be represented as $\hat{H}_{c,k}(t) \in \mathbb{C}$, where $c$ indexes the input channel, $k$ denotes the subcarrier index, and $t$ denotes the packet time index.

In essence, user motion, human breathing, body presence, and environmental dynamics perturb wireless propagation paths.
These perturbations change the temporal and spectral patterns of $\hat{H}_{c,k}(t)$, making CSI a useful signaling indicator for passive sensing and monitoring. Formally, the raw CSI is complex-valued:
\begin{equation}
\hat{H}_{c,k}(t)
=
A_{c,k}(t)e^{j\phi_{c,k}(t)},
\end{equation}
where $A_{c,k}(t)$ is the amplitude and $\phi_{c,k}(t)$ is the phase.
In practical commodity Wi-Fi systems, CSI phase can be unstable due to timing offsets, frequency synchronization errors, and hardware-dependent distortions. 
Therefore, it is common to use the amplitude component as the sensing model input:
\begin{equation}
A_{c,k}(t)
=
|\hat{H}_{c,k}(t)|.
\end{equation}
This amplitude-only design avoids relying on device-specific phase calibration and provides a robust input representation for commodity Wi-Fi sensing \cite{yang2023sensefi}\footnote{When reliably calibrated phase is available, our proposed framework can be extended by treating phase or complex-valued CSI components as additional input channels.}.
In this way, a sensing window is formed by aggregating CSI amplitudes over $T$ packet time indices and $K$ subcarriers.
The resulting temporal-spectral CSI tensor is defined as
\begin{equation}
X \in \mathbb{R}^{C \times K \times T},
\qquad
X_{c,k,t}=A_{c,k}(t).
\end{equation}
To reduce scale variations across devices and environments, each CSI window is independently standardized by subtracting its mean amplitude and dividing by its standard deviation.

This representation forms a natural temporal-spectral field.
The temporal axis captures motion evolution and physiological dynamics, while the subcarrier axis captures frequency-selective fading and multipath correlation.
Such structure motivates our temporal-spectral tokenization and channel variation-aware masking design as introduced in Sec.~\ref{subsec:channel_variation_masking}.

\begin{figure}[t]
    \centerline{\includegraphics[scale=0.33]{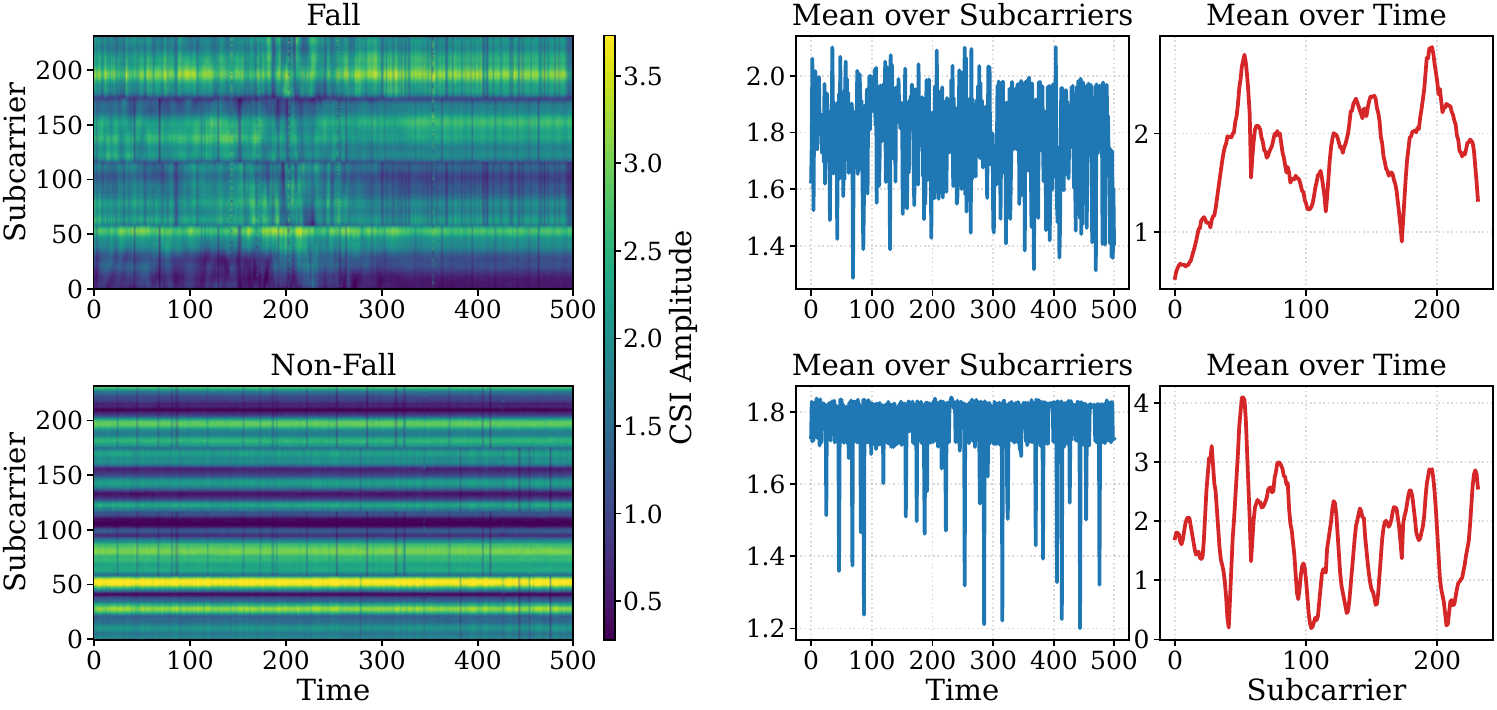}}
        \vspace{-0.5em}
    \caption{ Illustrative CSI amplitude examples for Fall and Non-Fall samples. 
    \textit{Left}: temporal-spectral CSI heatmaps. 
    \textit{Right}: aggregated temporal and subcarrier profiles obtained by averaging over subcarriers and time, respectively. 
    }
    \label{fig:csi_example}
    \vspace{-0.5em}
\end{figure}

\begin{figure*}[t]
    \centerline{\includegraphics[scale=0.93]{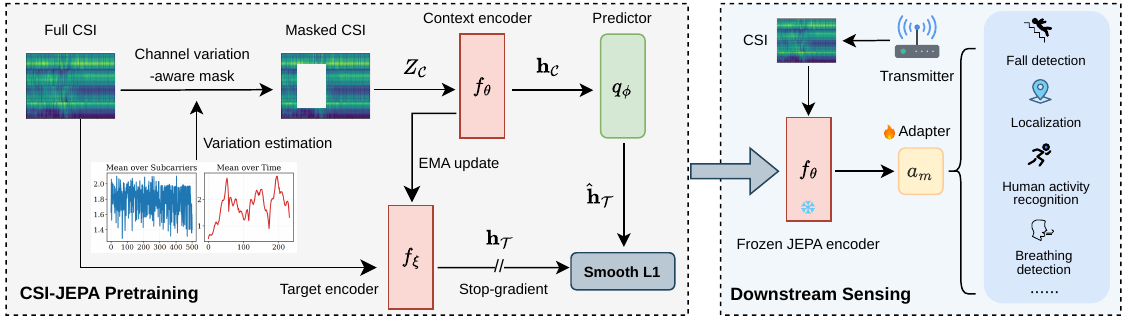}}
    
    \caption{{Overview of CSI-JEPA. The framework performs self-supervised predictive pretraining on temporal-spectral CSI samples using channel variation-aware masking, an online encoder, a predictor, and an EMA target encoder. After pretraining, the encoder is frozen and adapted to downstream Wi-Fi sensing tasks with lightweight task-specific adapters.}}
    \label{csijepa}
\end{figure*}

\textbf{Observations.} Fig.~\ref{fig:csi_example} illustrates representative CSI amplitude examples from a wireless human activity sensing task, specifically fall and non-fall events, where the goal is to distinguish abrupt, high-dynamic body motions from normal daily activities based on channel responses. 
The two classes exhibit visibly different temporal-spectral patterns in the CSI heatmaps. 
In addition, the aggregated temporal and subcarrier profiles show distinct structures, suggesting that both dimensions provide useful and complementary sensing cues. 
This observation motivates modeling CSI as a structured temporal-spectral field for the masking policy in JEPA that is aware of channel variations along both time and subcarrier horizons.

\subsection{Wi-Fi Sensing Task and Adaptation Objective}
\label{objective}

We consider a set of Wi-Fi sensing tasks $\mathcal{M}$.
Each task $m \in \mathcal{M}$ has a task-specific label space $\mathcal{Y}^{(m)}$. 
Given an input CSI tensor $X$, the objective is to predict
\begin{equation}
\hat{y}^{(m)}
=
\arg\max_{l \in \mathcal{Y}^{(m)}}
P(y^{(m)}=l \mid X).
\end{equation}
where $l$ denotes a candidate class for task $m$.

In this work, we evaluate each downstream task independently by training a separate lightweight adapter on top of a frozen pretrained encoder. 
This protocol isolates the quality of the learned CSI representation and avoids confounding the evaluation with task-balancing choices in  joint multi-task optimization.

Formally, our goal is to learn a reusable CSI encoder
$f_{\theta}: X \mapsto Z$
from unlabeled samples, such that downstream tasks can be adapted with only a small labeled subset. 
For task $m$, let $\mathcal{S}^{l}_{m}$ denote the labeled subset used for downstream adaptation. 
Given a frozen pretrained encoder $f_{\theta}$, we train a lightweight adapter $a_m$ by solving
\begin{equation}
\min_{a_m}
\mathbb{E}_{(X,y)\sim \mathcal{S}^{l}_{m}}
\left[
\mathcal{L}_{m}
\left(
a_m(f_{\theta}(X)), y
\right)
\right],
\quad
\mathrm{where}~\theta \ \mathrm{is~frozen}.
\end{equation}
where \(\mathcal{L}_{m}\) denotes the task-specific supervised loss for downstream task \(m\).

This objective reflects the practical setting where unlabeled CSI can be obtained from normal Wi-Fi operation, while labeled CSI is expensive to collect. 
It also matches our frozen-encoder evaluation protocol, where downstream performance reflects the quality and reusability of the pretrained CSI representation across sensing tasks and deployment domains.

\section{Proposed Channel Variation-Aware Predictive
Representation Learning}
To resolve the formulated problem in Sec.~\ref{objective}, we propose CSI-JEPA that consists of four components as shown in Fig.~\ref{csijepa}. 
First, a temporal-spectral tokenizer converts CSI samples into patch tokens while preserving the two-dimensional structure over time and subcarriers. 
Second, a channel variation-aware masking module estimates local channel variation from temporal and subcarrier-domain changes and selects the most informative target regions on the patch grid.
Third, an online encoder and a lightweight predictor infer the latent representations of masked target regions from visible context. 
Fourth, an exponential moving average (EMA) target encoder provides stable target representations for latent prediction.
During the pretraining stage, task labels are ignored and the model is optimized only by a predictive latent loss. 
After pretraining, the online encoder is frozen and transferred to downstream sensing tasks with lightweight task-specific adapters.

\subsection{Temporal-Spectral Tokenization}

CSI is not an unstructured vector. 
The temporal dimension reflects motion evolution and physiological dynamics, while the subcarrier dimension captures frequency-selective fading and multipath correlation. 
Therefore, we tokenize CSI along temporal and subcarrier dimensions jointly.

Given an input sample $X \in \mathbb{R}^{C \times K \times T}$, we divide the subcarrier-time plane into non-overlapping patches of size $P_K \times P_T$. 
This produces an $N_K \times N_T$ patch grid, where $N_K = K/P_K$ and $N_T = T/P_T$. 

Then, we use a convolutional patch embedder with kernel size and stride $(P_K, P_T)$ to extract non-overlapping temporal-spectral patch tokens:
\begin{equation}
Z_0
=
\mathrm{PatchEmbed}(X)
\in
\mathbb{R}^{N_K \times N_T \times D},
\end{equation}
where $D$ is the token embedding dimension.

To preserve subcarrier and temporal locations, we add fixed two-dimensional sine-cosine positional embeddings:
\begin{equation}
\widetilde{Z}_{i,j}
=
Z_{0,i,j}
+
E^{\mathrm{pos}}_{i,j},
\end{equation}
where $E^{\mathrm{pos}}_{i,j}$ denotes the fixed positional embedding for patch location $(i,j)$ on the subcarrier-time patch grid.
The resulting patch grid is then flattened into a token sequence
$Z=\{\tilde{z}_1,\tilde{z}_2,\ldots,\tilde{z}_N\}$, where $N=N_KN_T$.

\subsection{Channel Variation-Aware Masking}
\label{subsec:channel_variation_masking}

A key design question in CSI-based self-supervised learning is how to select masked target regions that are informative for representation learning while preserving sufficient global context. 
Uniform random block masking treats all temporal-spectral regions equally, but CSI measurements are not uniformly informative. 
Human motion and environmental changes often induce localized channel variations over both time and subcarrier horizons. 
In contrast, masking an entire temporal region or an entire subcarrier band can be overly destructive, since it may remove broad structural information that is needed for reliable learning.

As illustrated in Fig.~\ref{fig:csi_example}, informative sensing patterns exist along both temporal and subcarrier domains. 
This motivates a masking strategy that is aware of channel variations explicitly. 
We therefore introduce channel variation-aware temporal-spectral masking, which estimates a local channel-variation map from changes in CSI samples along the temporal and subcarrier dimensions and samples target blocks that cover regions with stronger channel dynamics.
Specifically, the temporal variation reflects time-varying channel properties induced by motion and Doppler effects, while subcarrier-domain variation reflects frequency-selective fading, multipath propagation, and interference patterns. 
This makes the predictive pretext task focus on sensing-informative  regions rather than arbitrary locations.

Given a normalized CSI tensor $X \in \mathbb{R}^{C \times K \times T}$, we first compute channel variation along the temporal dimension as
\begin{equation}
V_t(t,k)
=
\frac{1}{C}
\sum_{c=1}^{C}
\left|
X_{c,k,t} - X_{c,k,t-1}
\right|,
\end{equation}
and then derive the subcarrier-domain channel variation as
\begin{equation}
V_k(t,k)
=
\frac{1}{C}
\sum_{c=1}^{C}
\left|
X_{c,k,t} - X_{c,k-1,t}
\right|.
\end{equation}
For those boundary positions, we set the missing differences to zero. 
Here, $V_t$ captures temporal channel dynamics, while $V_k$ captures local subcarrier-domain variation. The variation scores are averaged across input channels when multiple channels are present.
In this way, the two variation maps are combined into a new channel-variation map:
\begin{equation}
M(t,k)
=
\lambda V_t(t,k)
+
(1-\lambda)V_k(t,k),
\end{equation}
where $\lambda \in [0,1]$ balances temporal and subcarrier-domain contributions. 
Unless otherwise specified, we set $\lambda=0.5$.

Since CSI-JEPA operates on temporal-spectral patches, we average
$M(t,k)$ within each patch and then within each feasible rectangular
target block to obtain its block-level variation score $R(a,b)$, where
$(a,b)$ denotes the upper-left corner of the block. Let $\Omega$ denote the set of all feasible block locations.
The target block location is then sampled according to the normalized block-level variation score:
\begin{equation}
p(a,b)
=
\frac{R(a,b)+\epsilon}
{\sum_{(u,v)\in\Omega}
\left(R(u,v)+\epsilon\right)},
\end{equation}
where $\epsilon$ is a small constant for numerical stability.

To avoid making the masking policy overly deterministic, we mix the variation-guided distribution with uniform block sampling:
\begin{equation}
p_{\mathrm{mask}}(a,b)
=
(1-\eta)p(a,b)
+
\eta\frac{1}{|\Omega|},
\end{equation}
where $\eta$ is the exploration probability. 
After sampling a target block, the visible context becomes
$\mathcal{C} = \mathcal{G} \setminus \mathcal{T}$,
where $\mathcal{T}$ denotes the target region and
$\mathcal{G}$ is the full patch grid.

This design preserves the advantages of block-based latent prediction while selecting masked regions according to local CSI variations.
It avoids the excessive information removal of full time-axis or full subcarrier-axis masking, and instead focuses the prediction task on locally informative regions where channel dynamics are strongest. 
As a result, the encoder is encouraged to model motion-sensitive temporal transitions jointly with correlated subcarrier-domain variations, which better matches the physical structure of CSI-based sensing than purely random or single-dimension masking.

\begin{algorithm}[t]
\caption{CSI-JEPA Pretraining}
\label{alg:csijepa}
\begin{algorithmic}[1]
\Require Unlabeled CSI set $\mathcal{U}$, online encoder $f_\theta$,
predictor $q_\phi$, target encoder $f_\xi$
\Ensure Updated $\theta$, $\phi$, and $\xi$
\For{each minibatch $\mathcal{B} \subset \mathcal{U}$}
\State $\mathcal{L}_{\mathrm{JEPA}} \gets 0$
    \For{each CSI window $X \in \mathcal{B}$}
        \State Normalize and tokenize $X$ into temporal--subcarrier patches $Z$
        \State Compute $V_t$, $V_k$, and $M=\lambda V_t+(1-\lambda)V_k$
        \State Compute scores \(R(a,b)\) and sample target block  $\mathcal{T}$
        \State $\mathcal{C} \gets \mathcal{G} \setminus \mathcal{T}$
        \Comment{Visible context}
        \State $h_{\mathcal{C}} \gets f_\theta(Z_{\mathcal{C}})$
        \State $h_{\mathcal{T}} \gets \operatorname{sg}(f_\xi(Z)_{\mathcal{T}})$
        \State $\hat{h}_{\mathcal{T}}
        \gets q_\phi(h_{\mathcal{C}}, E^{pos}_{\mathcal{T}})$
        \State $\mathcal{L}_{\mathrm{JEPA}}
        \gets \mathcal{L}_{\mathrm{JEPA}} +
        \operatorname{SmoothL1}(\hat{h}_{\mathcal{T}}, h_{\mathcal{T}})$
    \EndFor
    \State Update $\theta$ and $\phi$ using $\mathcal{L}_{\mathrm{JEPA}}$
    \State $\xi \gets \mu\xi + (1-\mu)\theta$
    \Comment{EMA target update}
\EndFor
\end{algorithmic}
\end{algorithm}

\subsection{Predictive Latent Pretraining}

Given the tokenized CSI sample $Z$ and the target region $\mathcal{T}$ sampled by the channel variation-aware masking, CSI-JEPA learns by predicting the latent representations of masked target patches from visible context patches. 
The visible context is denoted by
$Z_{\mathcal{C}} = \{\tilde{z}_i \mid i \in \mathcal{C}\}$,
where $\mathcal{C}$ denotes the visible context patch set.

\noindent\textbf{Online and target encoders.}
The online encoder maps visible context tokens into latent representations:
\begin{equation}
\mathbf{h}_{\mathcal{C}} = f_{\theta}(Z_{\mathcal{C}}).
\end{equation}
This encoder can be instantiated by any temporal-spectral backbone. 
In this work, we use a lightweight Vision Transformer (ViT)-style encoder~\cite{dosovitskiy2020image} to model long-range dependencies among visible temporal-spectral tokens through multi-head self-attention.

On the other hand, the target encoder has the same architecture as the online encoder but uses parameters $\xi$ updated by EMA:
\begin{equation}
\xi \leftarrow \mu \xi + (1-\mu)\theta,
\end{equation}
where $\mu$ is the momentum coefficient. 
Given the full token sequence $Z$, the target encoder produces stop-gradient latent targets for the masked region:
\begin{equation}
\mathbf{h}_{\mathcal{T}}
=
\mathrm{sg}
\left(
f_{\xi}(Z)_{\mathcal{T}}
\right),
\end{equation}
where $f_{\xi}(Z)_{\mathcal{T}}$ denotes the target encoder outputs indexed by the masked target set $\mathcal{T}$, and $\mathrm{sg}(\cdot)$ denotes the stop-gradient operation.

\noindent\textbf{Predictor.}
The predictor maps the context representation into the target latent space. 
Since target tokens are not visible to the online encoder, the predictor receives the context representation and the positional information of target patches:
\begin{equation}
\hat{\mathbf{h}}_{\mathcal{T}}
=
q_{\phi}(\mathbf{h}_{\mathcal{C}}, E^{\mathrm{pos}}_{\mathcal{T}}),
\end{equation}
where $E^{\mathrm{pos}}_{\mathcal{T}}$ denotes the positional embeddings of the target patches. 
The predictor is intentionally lightweight compared with the encoder, so that the encoder is encouraged to learn reusable CSI representations instead of relying on a high-capacity prediction head.

\noindent\textbf{Learning objective.}
CSI-JEPA minimizes the discrepancy between the predicted target representations and the EMA target representations:
\begin{equation}
\mathcal{L}_{\mathrm{JEPA}}
=
\frac{1}{|\mathcal{T}|}
\sum_{i \in \mathcal{T}}
\ell
\left(
\hat{h}_i,
\mathrm{sg}(h_i)
\right),
\end{equation}
where $\hat{h}_i$ is the predicted target embedding, $h_i$ is the corresponding EMA target embedding, $\mathrm{sg}(\cdot)$ denotes stop-gradient, and $\ell(\cdot,\cdot)$ is the Smooth $L_1$ loss.
For a predicted target embedding $\hat{h}_i$ and its stop-gradient target embedding $\mathrm{sg}(h_i)$, the loss is defined element-wise as
\begin{equation}
\ell
\left(
\hat{h}_i,
\mathrm{sg}(h_i)
\right)
=
\frac{1}{D}
\sum_{j=1}^{D}
\begin{cases}
\frac{1}{2} r_{i,j}^{2},
&
|r_{i,j}| < 1,
\\
|r_{i,j}| - \frac{1}{2},
&
\mathrm{otherwise},
\end{cases}
\end{equation}
where $D$ is the embedding dimension and
\begin{equation}
r_{i,j} = \hat{h}_{i,j} - \mathrm{sg}(h_{i,j}).
\end{equation}
The online encoder parameters $\theta$ and predictor parameters $\phi$ are updated by backpropagation, while the target encoder parameters $\xi$ are updated only through EMA.

We pretrain CSI-JEPA on heterogeneous unlabeled CSI samples from multiple sensing tasks:
\begin{equation}
\mathcal{U}
=
\bigcup_{m \in \mathcal{M}}
\mathcal{U}^{(m)}.
\end{equation}
No task labels are needed during pretraining, and the same JEPA objective is applied to all unlabeled CSI samples.
This design encourages the encoder to learn task-agnostic CSI primitives, including temporal continuity, activity periodicity, cross-subcarrier consistency, and other reusable patterns shared across sensing objectives.
Algorithm 1 summarizes the complete CSI-JEPA pretraining procedure.

\subsection{Downstream Adaptation}

After the self-supervised pretraining stage, we transfer the online encoder $f_{\theta}$ to downstream Wi-Fi sensing tasks.
To directly evaluate the quality of the learned representation and reduce the cost of task adaptation, we freeze the pretrained encoder for all downstream evaluations.
Only a lightweight task-specific adapter is trained using a limited number of labeled samples for each task.

For a labeled CSI sample $X$, the frozen encoder produces token representations
$\mathbf{h}=f_{\theta}(X)=\{h_1,h_2,\ldots,h_N\}$.
Since each downstream label is assigned to the entire CSI window, we aggregate token representations into a sample-level embedding by average pooling:
\begin{equation}
r
=
\frac{1}{N}
\sum_{i=1}^{N}
h_i.
\end{equation}

For each downstream task $m$, we train a separate adapter $a_m$ on top of the frozen representation using the labeled subset $\mathcal{S}_{m}^{l}$:
\begin{equation}
\min_{a_m}
\mathbb{E}_{(X,y) \sim \mathcal{S}_{m}^{l}}
\left[
\mathcal{L}_{\mathrm{CE}}
\left(
a_m(r), y
\right)
\right],
\qquad
\theta \ \mathrm{frozen}.
\end{equation}
This independent adaptation protocol is used for all downstream tasks to isolate the quality of the pretrained CSI representation.

We evaluate two frozen-encoder adaptation settings.
The first setting uses a linear classifier on top of the mean-pooled representation, which measures whether the pretrained features are linearly separable for downstream sensing tasks.
The second setting uses a lightweight two-layer MLP adapter, which evaluates whether task-relevant sensing information can be extracted by a small nonlinear head without updating the encoder.
For both settings, we vary the number of labeled samples used to train the adapter.
This allows us to evaluate label efficiency under limited supervision while keeping the pretrained encoder fixed.
Compared with full fine-tuning, this frozen-encoder protocol is computationally efficient and more directly reflects the reusability and transferability of the learned CSI representation as reported in Sec.~\ref{results}.

\section{Experimental Methodology}

\subsection{Benchmark and Tasks}
\label{subsec:experimental_setup}

\begin{table}[t]
\centering
\caption{Summary of the seven CSI-Bench sensing tasks.}
\small
\label{tab:task_summary}
\begin{tabular}{lcc}
\toprule
\textbf{Task} & \textbf{Type} & \textbf{\# Classes} \\
\midrule
Fall Detection (Fall) & Individual & 2 \\
Motion Source Recognition (MSR) & Individual & 2 \\
Breathing Detection (Breath) & Individual & 2 \\
Room-Level Localization (Loc) & Individual & 6 \\
Human Activity Recognition (HAR) & Shared & 6 \\
Proximity Recognition (PR) & Shared & 4 \\
User Identification (UID) & Shared & 6 \\
\bottomrule
\end{tabular}
\vspace{-2em}
\end{table}

We evaluate CSI-JEPA on seven CSI-Bench Wi-Fi sensing tasks~\cite{zhu2025csi}, including four individually defined tasks and three shared-data tasks as shown in Table~\ref{tab:task_summary}. 
CSI-Bench is a large-scale, in-the-wild Wi-Fi sensing benchmark collected from commercial Wi-Fi routers and IoT edge devices across diverse users, hardware platforms, and indoor environments. It contains over 461 hours of CSI recordings from 35 users, 26 environments, and 16 device configurations. 
This provides a realistic testbed for evaluating reusable CSI representations across heterogeneous sensing tasks and deployment settings.

Following the CSI-Bench preprocessing protocol, CSI from different antenna links is organized along a combined link--subcarrier feature axis. For example, a \(2\times2\) MIMO configuration with four spatial links and 58 subcarriers per link yields 232 link--subcarrier features. We normalize each sample and clip or zero-pad the temporal and feature dimensions to obtain a fixed \(500\times232\) input. The resulting tensor contains a single amplitude channel, while antenna-link information is represented along the feature dimension rather than averaged across antennas.

For self-supervised pretraining, we adopt only valid unlabeled CSI samples from the training splits of the seven tasks, resulting in approximately 151K CSI windows. 
No validation or test samples are used during pretraining, ensuring strict separation between pretraining and downstream evaluation.
For downstream adaptation, we use label budgets of $\{10,100,500,1000,B_{\max}\}$, where $B_{\max}$ denotes the task-specific maximum training budget. Labeled samples are drawn only from the corresponding training split, while validation data are used for model selection and accuracy and weighted F1-score are reported on the held-out test split. Because CSI-Bench has imbalanced task labels and different training-set sizes, we use the full training split when it contains fewer than 10K samples and otherwise cap the maximum budget at 10K labels. 

\subsection{Implementation Details}
All methods use amplitude-only CSI inputs. Each CSI sample is independently normalized and standardized to a fixed input size. The model architecture is summarized in Table~\ref{tab:jepa_architecture}. We pretrain CSI-JEPA for 20 epochs using AdamW with learning rate $10^{-4}$ and weight decay $10^{-5}$. The target encoder is updated by exponential moving average, with the momentum linearly increased from 0.996 to 1.0 during pretraining. The exploration probability in channel variation-aware masking is set to $\eta=0.3$. We save checkpoints during pretraining and use the 15-epoch checkpoint in the main experiments, since downstream performance becomes stable after a few pretraining epochs in our validation study. For downstream adaptation, all adapters are trained with AdamW and batch size 32 for at most 20 epochs, with early stopping based on validation accuracy.

\begin{table}[t]
\centering
\caption{Implementation details of CSI-JEPA.}
\label{tab:jepa_architecture}
\small
% \footnotesize
\setlength{\tabcolsep}{4pt}
\renewcommand{\arraystretch}{1.05}
\begin{tabular}{p{0.27\columnwidth} p{0.65\columnwidth}}
\toprule
\textbf{Module} & \textbf{Configuration} \\
\midrule
Input CSI & Amplitude CSI tensor with input size $C=1$, $K=232$, and $T=500$. \\

Patch embedding & 2D convolutional patch embedder with patch size $P_K \times P_T = 8 \times 25$, token dimension 256.\\

Positional encoding & Fixed 2D sine-cosine positional encoding added to temporal-spectral patch tokens. \\

Online encoder & ViT-style Transformer with 6 blocks, hidden dimension 256, and 8 attention heads. \\

Target encoder & EMA copy of the online encoder with the same architecture. \\

Predictor & Lightweight Transformer with 3 blocks, hidden dimension 192, and 6 attention heads. \\

Prediction head & Linear projection mapping predictor outputs back to the 256-dimensional target space. \\

Pretraining loss & Smooth L1 latent regression between predicted target tokens and EMA target tokens. \\

Masking & One contiguous rectangular target block selected based on channel variation scores, with block area sampled from 15\%--30\% of all patches and aspect ratio sampled from [0.5, 2.0]. \\

Linear head & Mean pooling followed by LayerNorm and a linear probing head.\\

MLP head & Mean pooling followed by LayerNorm and a 2-layer MLP probing head.\\

\bottomrule

\end{tabular}
\vspace{-1em}
\end{table}

\subsection{Baselines and Variants}
We evaluate CSI-JEPA from three perspectives: supervised baselines, an alternative SSL baseline, and CSI-JEPA design variants for adaptation, backbone, and masking ablations.

\noindent\textbf{1) Supervised baselines.}
We compare CSI-JEPA with three models trained directly on labeled CSI data:
\begin{itemize}
    \item \textbf{Raw linear~\cite{rosenblatt1958perceptron}:} a linear classifier trained on flattened CSI amplitude features.
    \item \textbf{Raw MLP~\cite{rumelhart1986learning}:} a two-layer MLP trained from scratch on flattened CSI amplitude features.
    \item \textbf{Sup. Transformer~\cite{vaswani2017attention}:} a supervised Transformer encoder trained end-to-end from labeled CSI samples.
\end{itemize}
These baselines evaluate whether self-supervised pretraining can improve downstream sensing performance compared with direct supervised training, especially under limited label budgets.

\noindent\textbf{2) SSL baseline based on CSI-MAE~\cite{jiang2026csi}.}
We implement a CSI-MAE-style masked reconstruction baseline to compare CSI-JEPA with reconstruction-based SSL. 
For a fair comparison, this baseline uses the same temporal-subcarrier tokenizer, patch size, ViT encoder scale, unlabeled pretraining corpus, and pretraining budget as CSI-JEPA, but replaces latent target prediction with masked CSI amplitude reconstruction.
Detailed adaptation of CSI-MAE to our Wi-Fi sensing task setting is described in Sec.~\ref{sec:mae_comparison}.

\begin{figure*}[t]
    \centering
    \begin{subfigure}{0.9\textwidth}
        \centering
        \includegraphics[width=\textwidth]{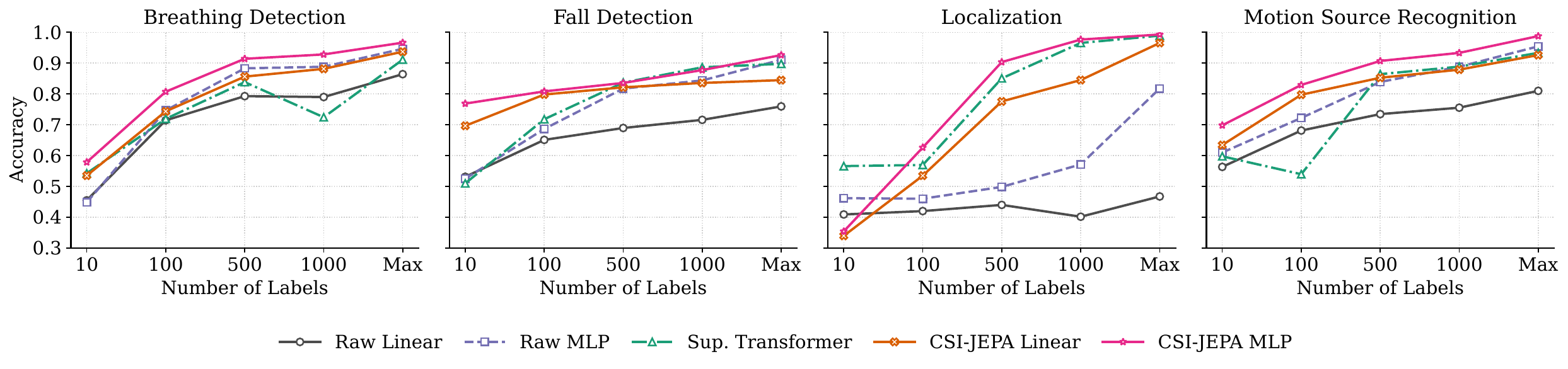}
        % \caption{Accuracy.}
        \label{fig:single_task_acc}
    \end{subfigure}

    \vspace{-0.7em}

    \begin{subfigure}{0.9\textwidth}
        \centering
        \includegraphics[width=\textwidth]{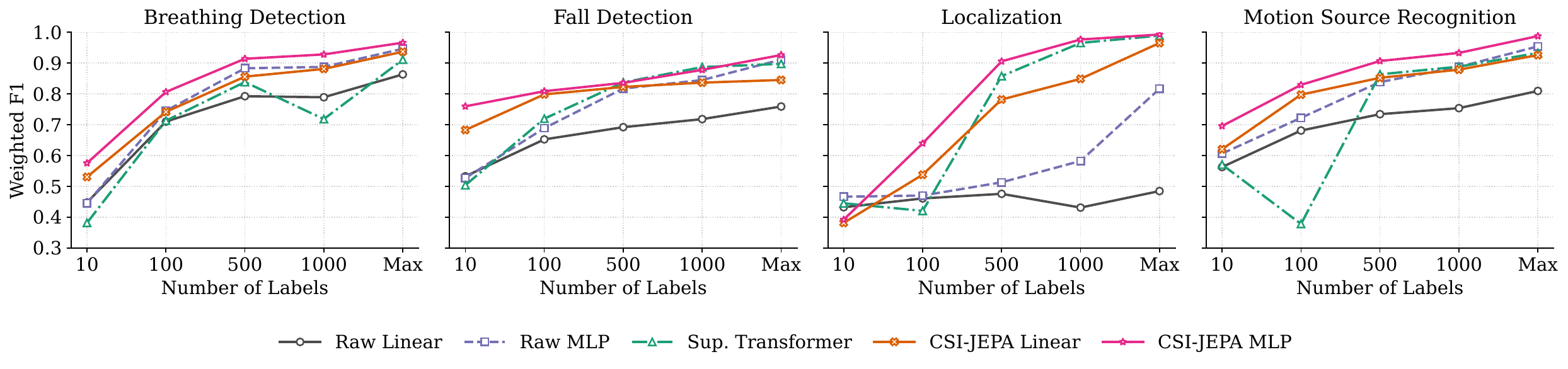}
        % \caption{Weighted F1-score.}
        \label{fig:single_task_f1}
    \end{subfigure}
    \vspace{-2.5em}

    \caption{Accuracy and weighted F1-score under different label budgets on four individually defined CSI sensing tasks.}
    \label{fig:single_task_results}
    \vspace{-0.5em}
\end{figure*}

\noindent\textbf{3) CSI-JEPA adaptation variants.}
After pretraining, the CSI-JEPA encoder is frozen for all downstream tasks. 
We evaluate two lightweight task-specific adapters:
\begin{itemize}
    \item \textbf{CSI-JEPA linear probe:} a linear classifier trained on the mean-pooled frozen representation.
    \item \textbf{CSI-JEPA MLP probe:} a lightweight MLP adapter trained on the mean-pooled frozen representation.
\end{itemize}

\noindent\textbf{4) CSI-JEPA backbone variants.}
To study the effect of encoder architecture, we compare three backbone variants under the same JEPA pretraining protocol:
\begin{itemize}
    \item \textbf{ViT backbone~\cite{dosovitskiy2020image}:} directly tokenizes CSI amplitude patches and processes them with a Transformer encoder.
    \item \textbf{CNN backbone~\cite{726791}:} uses convolutional layers to extract local temporal-subcarrier features from CSI heatmaps.
    \item \textbf{CNN-ViT backbone:} first applies a CNN front-end for local feature extraction, followed by a Transformer encoder for global temporal-subcarrier modeling.
\end{itemize}
This comparison evaluates whether local convolutional inductive bias, global self-attention, or their combination is more effective for JEPA-based CSI representation learning.

\noindent\textbf{5) Masking variants.}
To study the effect of target selection during pretraining, we compare four masking strategies:
\begin{itemize}
    \item \textbf{Time-based masking:} masks a contiguous temporal region across all subcarriers.
    \item \textbf{Subcarrier-based masking:} masks a contiguous subcarrier band across all time steps.
    \item \textbf{Temporal-spectral masking:} uniformly samples a contiguous rectangular block on the temporal-subcarrier patch grid.
    \item \textbf{Channel variation-aware masking:} samples a rectangular target block according to channel variation scores.
\end{itemize}

\begin{table}[t]
\centering
\caption{Domain shift from CSI-Bench pretraining to MMFi adaptation.}
\label{tab:transfer_shift}
\small
\begin{tabular}{lll}
\toprule
Aspect & CSI-Bench~\cite{zhu2025csi} & MMFi~\cite{yang2023mm} \\
\midrule
Task supervision & None & 2 unseen target tasks \\
Environment & Source environments & Unseen environments \\
Hardware & 16 device configs. & Different Wi-Fi hardware \\
CSI organization & $1\times500\times232$ & $3\times114\times10$ per frame \\
Encoder stage & Pretrained & Frozen transfer \\
Target labels & 0 & 1,000 / 240 \\
\bottomrule
\end{tabular}
\vspace{-1.3em}
\end{table}

\subsection{Cross-Dataset Transfer Setup}
\label{sec:cross_dataset_setup}

To evaluate representation transfer under substantial domain shift, we use MMFi \cite{yang2023mm} as an independent target dataset. As summarized in Table~\ref{tab:transfer_shift}, compared with CSI-Bench, MMFi introduces previously unseen sensing tasks, environments, and hardware, and even organizes CSI measurements differently. The CSI-JEPA encoder is pretrained exclusively on unlabeled CSI-Bench data and remains frozen throughout MMFi adaptation; only the downstream classification head is trained using labeled MMFi samples. Each MMFi CSI amplitude frame has shape $3\times114\times10$, corresponding to three antenna links, 114 subcarriers, and 10 temporal samples. We flatten the antenna-link and subcarrier dimensions into 342 features, concatenate consecutive frames to form a 500-sample temporal window, linearly resize the feature dimension from 342 to 232, and standardize each window to obtain the $1\times500\times232$ input expected by CSI-JEPA.

We construct two binary target tasks from MMFi action labels. \emph{Motion Intensity} groups selected actions into low-motion and high-motion classes based on their action categories, while \emph{Bow vs.\ Jump} distinguishes bowing from jumping. Motion Intensity uses 1,000 labeled training windows and Bow vs.\ Jump uses 240. We use a fixed subject-disjoint split for all methods. For CSI-JEPA, mean pooling over the frozen encoder tokens produces a 256-dimensional representation that is passed to either a linear or two-layer MLP head. For the raw-CSI control, the identically preprocessed $500\times232$ input is flattened and reduced to 256 dimensions using parameter-free adaptive average pooling before applying the same downstream head. Thus, paired methods use identical trainable head architectures and parameter counts. All results are averaged over three random seeds.

\section{Evaluation Results}
\label{results}
\subsection{Label-Efficient Downstream Adaptation}

First, Fig.~\ref{fig:single_task_results} reports the accuracy and weighted F1-score on four CSI-based sensing tasks under different label budgets. Across all tasks, CSI-JEPA consistently improves over the raw-feature baselines, showing that predictive pretraining learns reusable CSI representations beyond directly fitting classifiers on flattened CSI amplitudes. The advantage is especially clear under \textit{limited} supervision. With only 100 or 500 labeled samples, CSI-JEPA with an MLP probe already achieves strong performance on all four sensing tasks.
\begin{table}[t]
\centering
\caption{Label-efficiency summary of CSI-JEPA MLP over the supervised Transformer on four single-task settings.}
\label{tab:label_efficiency_summary}
\small
{
\begin{tabular}{lcccccc}
\toprule
\multirow{2}{*}{Task}
& \multicolumn{2}{c}{\(\Delta\)Acc}
& \multicolumn{2}{c}{\(\Delta\)F1}
& \multicolumn{2}{c}{Label Saving} \\
\cmidrule(lr){2-3}
\cmidrule(lr){4-5}
\cmidrule(lr){6-7}
& Mean & Max
& Mean & Max
& Saving
& Budget Pair \\
\midrule
Breath
& 9.18 & 20.38
& 12.55 & 20.98
& 95.0\%
& $500 \rightarrow 10{,}000$ \\

Fall
& 7.35 & 25.89
& 7.26 & 25.59
& 90.0\%
& $10 \rightarrow 100$ \\

Loc
& -1.75 & 5.66
& 4.57 & 21.92
& 0.0\%
& -- \\

MSR
& 10.64 & 28.98
& 14.38 & 45.13
& 80.0\%
& $100 \rightarrow 500$ \\
\bottomrule
\end{tabular}
}
% \vspace{-1.5em}
\end{table}

\begin{table*}[t]
\centering
\caption{Cross-dataset transfer from CSI-Bench to MMFi.
\normalfont 
Results are mean $\pm$ standard deviation over three random seeds (\%). Gains are measured against the corresponding raw CSI baseline with the same downstream head.}
\label{tab:mmfi_transfer}
\small
\begin{tabular}{llccccc}
\toprule
\textbf{MMFi Task} &
\textbf{Adapter} &
\textbf{Method} &
\textbf{Accuracy} &
\textbf{Acc. Gain} &
\textbf{Weighted F1} &
\textbf{Balanced Acc.} \\
\midrule

\multirow{4}{*}{\begin{tabular}[c]{@{}l@{}}Motion Intensity\\(1,000 labels)\end{tabular}}
& \multirow{2}{*}{Linear}
& No Pretraining
& $50.10 \pm 4.03$
& -- 
& $52.23 \pm 3.74$
& $49.10 \pm 2.47$ \\
&
& \textbf{CSI-JEPA}
& $\mathbf{70.59 \pm 0.92}$
& $\mathbf{+20.49}$
& $\mathbf{70.80 \pm 0.89}$
& $\mathbf{65.36 \pm 1.07}$ \\
\cmidrule(lr){2-7}

& \multirow{2}{*}{MLP}
& No Pretraining
& $53.24 \pm 2.17$
& --
& $55.11 \pm 1.94$
& $50.74 \pm 2.04$ \\
&
& \textbf{CSI-JEPA}
& $\mathbf{68.92 \pm 2.14}$
& $\mathbf{+15.69}$
& $\mathbf{69.33 \pm 1.77}$
& $\mathbf{64.08 \pm 1.22}$ \\

\midrule

\multirow{4}{*}{\begin{tabular}[c]{@{}l@{}}Bow vs. Jump\\(240 labels)\end{tabular}}
& \multirow{2}{*}{Linear}
& No Pretraining
& $52.50 \pm 4.51$
& --
& $52.38 \pm 4.50$
& $52.50 \pm 4.51$ \\
&
& \textbf{CSI-JEPA}
& $\mathbf{71.67 \pm 3.82}$
& $\mathbf{+19.17}$
& $\mathbf{70.81 \pm 4.24}$
& $\mathbf{71.67 \pm 3.82}$ \\
\cmidrule(lr){2-7}

& \multirow{2}{*}{MLP}
& No Pretraining
& $56.67 \pm 4.02$
& --
& $56.51 \pm 4.17$
& $56.67 \pm 4.02$ \\
&
& \textbf{CSI-JEPA}
& $\mathbf{79.58 \pm 1.44}$
& $\mathbf{+22.92}$
& $\mathbf{79.42 \pm 1.47}$
& $\mathbf{79.58 \pm 1.44}$ \\

\bottomrule
\end{tabular}
\end{table*}

Second, compared with the supervised Transformer trained from scratch, CSI-JEPA shows more stable performance across label budgets. 
For example, on Breathing Detection and Motion Source Recognition tasks, the supervised Transformer exhibits noticeable fluctuations when the labeled budget is small, whereas CSI-JEPA improves more smoothly as the number of labels increases. 
This suggests that self-supervised predictive pretraining provides a stronger representation for downstream sensing than directly training a Transformer from limited labeled CSI data.
Among the two CSI-JEPA adaptation variants, the MLP probe generally performs best, 
indicating that the frozen representation contains useful sensing information that can be effectively extracted by a lightweight adapter. 
We also observe that even CSI-JEPA Linear adapter is competitive, which further suggests that the learned representation is largely separable for downstream sensing tasks even without updating the backbone.

\begin{figure}[t]
    \centering
    \begin{subfigure}{0.47\textwidth}
        \centering
        \includegraphics[width=\textwidth]{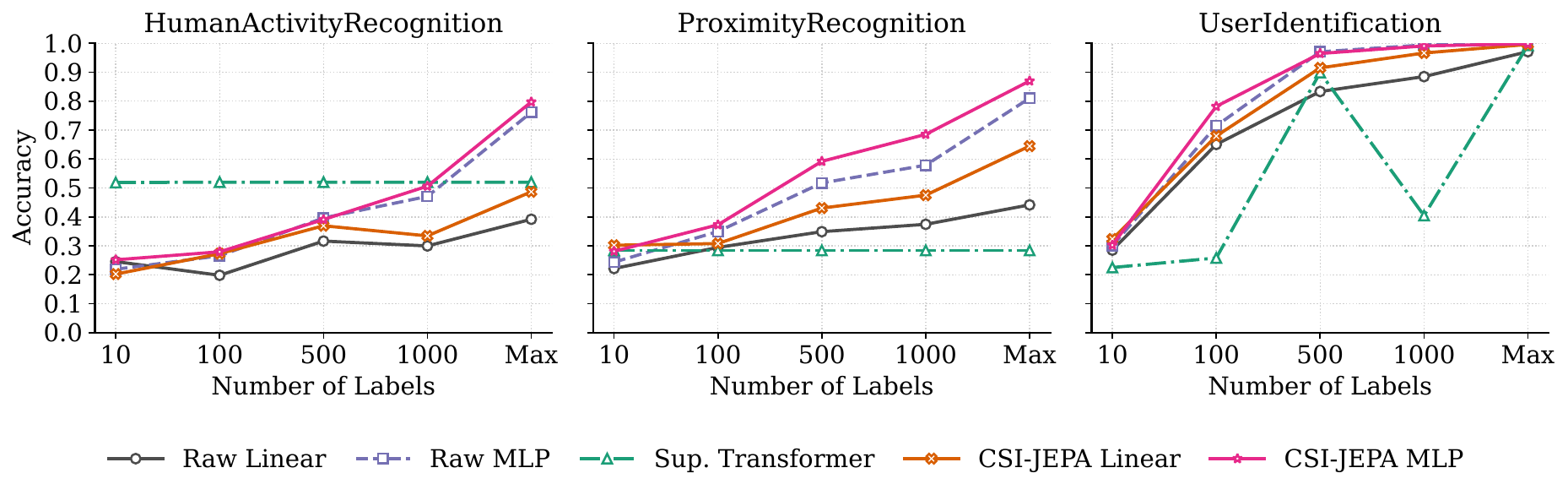}
        % \caption{Accuracy.}
    \end{subfigure}

    \begin{subfigure}{0.47\textwidth}
        \centering
        \includegraphics[width=\textwidth]{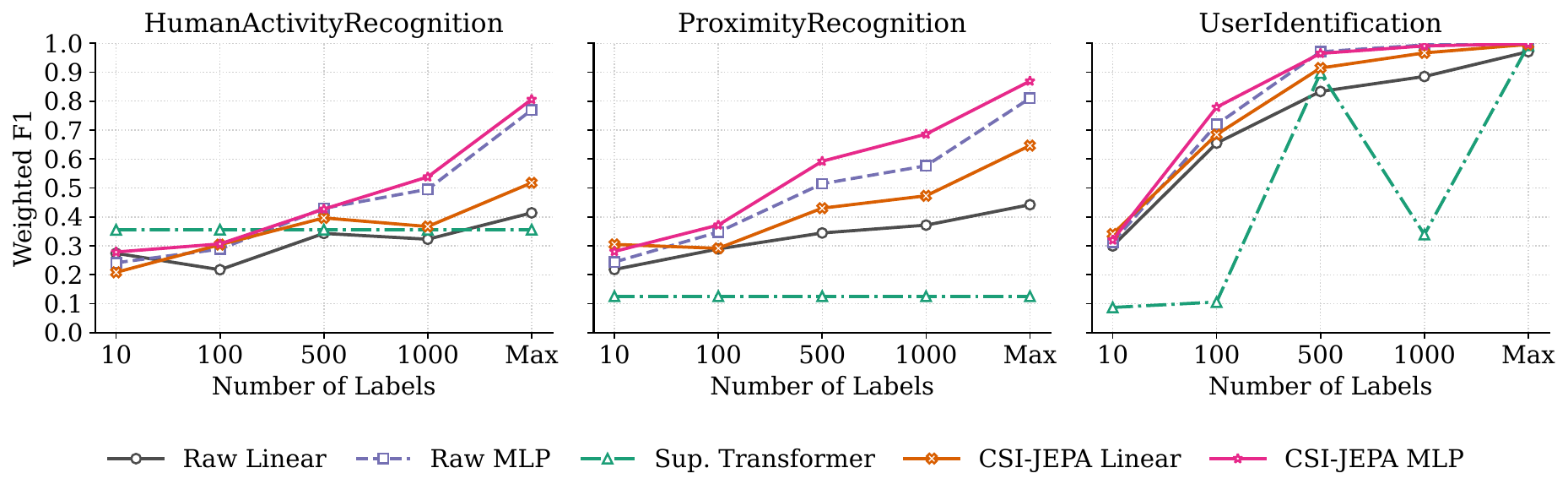}
        % \caption{Weighted F1-score.}
    \end{subfigure}
    \vspace{-1em}
    \caption{Accuracy and weighted F1-score under different label budgets on three shared-data tasks.}
    \label{multi_task_results}
    % \vspace{-1em}
\end{figure}

To further quantify the trends in Fig.~\ref{fig:single_task_results}, Table~\ref{tab:label_efficiency_summary} summarizes the label-efficiency gains of CSI-JEPA MLP over the supervised Transformer. We use the supervised Transformer as the reference baseline because it is the strongest supervised baseline, as validated in Fig.~\ref{fig:single_task_results}. Accuracy and F1 gains ($\Delta$) are computed at the same labeling budget and reported in percentage points (pp).
For label saving, we consider all Transformer operating points along the label-budget curve. For each operating point, we identify a matched Transformer--CSI-JEPA budget pair, where CSI-JEPA is allowed to perform at most 5 pp below the matched Transformer in accuracy. The 5-pp tolerance is one-sided and only limits underperformance by CSI-JEPA. Among all valid pairs, we report the pair with the largest positive label saving rate, computed as
\[
\mathrm{SavingRate}
=
1-\frac{b_J}{b_T},
\]
where $b_J$ and $b_T$ denote the CSI-JEPA and Transformer label budgets, respectively. If no positive-saving pair satisfies the matching criterion, the saving rate is reported as zero.

As shown in Table~\ref{tab:label_efficiency_summary}, CSI-JEPA achieves substantial same-budget gains over the supervised Transformer, particularly on Motion Source Recognition and Breathing Detection. CSI-JEPA improves mean accuracy by 10.64 pp on MSR and 9.18 pp on Breathing Detection, with corresponding mean F1 gains of 14.38 pp and 12.55 pp. The maximum F1 gain reaches 45.13 pp on MSR.

The matched-budget analysis further shows that CSI-JEPA can achieve comparable or better accuracy with substantially fewer labeled samples. On Breathing Detection, CSI-JEPA MLP with only \textbf{500} labels is matched to the supervised Transformer with \textbf{10,000} labels, corresponding to a saving rate of 95.0\%. On Fall Detection, CSI-JEPA MLP with only \textbf{10} labels is matched to the Transformer with \textbf{100} labels, yielding a 90.0\% saving rate. On Motion Source Recognition, CSI-JEPA MLP with \textbf{100} labels is matched to the Transformer with \textbf{500} labels, yielding an 80.0\% saving rate. For Localization, no positive label saving is observed under this criterion, although CSI-JEPA still provides positive mean and maximum F1 gains. 
Unlike activity-oriented sensing tasks, which often rely on transferable temporal patterns of channel variations, localization typically requires distinguishing fine-grained, location-dependent spatial signatures that are highly sensitive to environmental geometry and multipath propagation. Such spatial characteristics may not be fully captured by general-purpose representations learned through predictive pretraining, making task-specific labeled calibration more important. 
Overall, CSI-JEPA provides the largest label-efficiency benefits on multiple tasks where supervised learning remains data limited.

\begin{table*}[t]
\centering
\caption{Comparison of different masking strategies across seven tasks. 
\normalfont Results are reported as Acc/F1 (\%) using the MLP evaluation head. Bold marks a best result with $>1$ pp margin, and underline marks a best result with $\leq1$ pp margin. }
\label{tab:mask_comparison}
\small
\begin{tabular}{lcccccccccccccc}
\toprule
\multirow{2}{*}{\textbf{Mask}} 
& \multicolumn{2}{c}{\textbf{Fall}} 
& \multicolumn{2}{c}{\textbf{Loc}} 
& \multicolumn{2}{c}{\textbf{MSR}} 
& \multicolumn{2}{c}{\textbf{Breath}}
& \multicolumn{2}{c}{\textbf{HAR}}
& \multicolumn{2}{c}{\textbf{PR}}
& \multicolumn{2}{c}{\textbf{UID}} \\
\cmidrule(lr){2-3}
\cmidrule(lr){4-5}
\cmidrule(lr){6-7}
\cmidrule(lr){8-9}
\cmidrule(lr){10-11}
\cmidrule(lr){12-13}
\cmidrule(lr){14-15}
& Acc & F1
& Acc & F1
& Acc & F1
& Acc & F1
& Acc & F1
& Acc & F1
& Acc & F1 \\
\midrule
Time-based
& 88.93 & 88.92
& 97.99 & 97.94
& 98.67 & 98.67
& 96.74 & 96.74
& 77.29 & 78.30
& 84.26 & 84.16
& \underline{99.94} & \underline{99.94} \\

Subcarrier-based
& 83.96 & 84.09
& \underline{100.00} & \underline{100.00}
& 98.48 & 98.48
& 96.35 & 96.34
& 64.09 & 66.80
& 72.51 & 72.26
& 99.54 & 99.54 \\

Temporal-spectral
& 91.27 & 91.30
& \underline{100.00} & \underline{100.00}
& 98.41 & 98.41
& \underline{96.87} & \underline{96.87}
& 76.35 & 77.43
& 81.60 & 81.50
& 99.88 & 99.88 \\

\textbf{Channel-aware} (Ours)
& \textbf{92.59} & \textbf{92.62}
& 99.27 & 99.27
& \underline{98.75} & \underline{98.75}
& 96.62 & 96.62
& \textbf{79.74} & \textbf{80.60}
& \textbf{86.98} & \textbf{86.93}
& 99.83 & 99.83 \\
\bottomrule
\end{tabular}
% \vspace{-1em}
\end{table*}

Fig.~\ref{multi_task_results} further evaluates CSI-JEPA on three shared-data CSI sensing tasks. 
The results show that CSI-JEPA MLP consistently achieves the best or near-best performance across different label budgets.  
Particularly on Proximity Recognition, CSI-JEPA provides clear gains in the low- and medium-label regimes compared with raw-feature baselines, showing that the pretrained representation remains effective across diverse downstream sensing objectives.

\subsection{Generalization Beyond the Source Domain}
\label{sec:mmfi_transfer}

We next evaluate the cross-dataset transferability of CSI-JEPA representations by transferring the pretrained encoder beyond CSI-Bench to an unseen Wi-Fi sensing dataset. The CSI-JEPA encoder is pretrained only on unlabeled CSI-Bench samples and then transferred to MMFi~\cite{yang2023mm}, where we keep the encoder frozen and train only lightweight linear or MLP heads using limited labeled MMFi data. This setting introduces
substantial shifts in sensing hardware, CSI organization, and
downstream task definition.

Table~\ref{tab:mmfi_transfer} reports the results over three random seeds. CSI-JEPA consistently outperforms the corresponding raw CSI baseline under both lightweight linear and MLP adaptation. The largest gain is observed on Bow-vs.-Jump classification, where CSI-JEPA with an MLP head improves accuracy by \textbf{22.92 pp}, reaching $79.58\%$ with only 240 labeled samples. Notably, even with a linear
probe, CSI-JEPA achieves a \textbf{20.49 pp} accuracy improvement on motion-intensity classification, indicating that the transferred representation itself is substantially more discriminative than the raw CSI counterpart.
For the motion-intensity task, balanced accuracy provides a more informative measure than raw accuracy. CSI-JEPA
also achieves clear gains under this metric, confirming that the improvement is not explained by majority-class prediction.
Overall, these results provide direct evidence that the representation learned from CSI-Bench remains useful for adapting to sensing tasks from an independent CSI dataset without updating the pretrained encoder.

\subsection{Comparison with Reconstruction-Based SSL}
\label{sec:mae_comparison}

To compare CSI-JEPA with masked reconstruction-based self-supervised learning, we implement a CSI-MAE-style baseline from~\cite{jiang2026csi} within our benchmark setting. 
Specifically, we preserve its MAE-style masked reconstruction objective, use uniform random masking with a 75\% masking ratio, and include a learnable \texttt{[CLS]} token in both the encoder and decoder. 
For downstream evaluation, we also follow the original CSI-MAE adaptation protocol by using the encoder \texttt{[CLS]} representation with a linear prediction head. 
Since the original CSI-MAE formulation addresses positioning as a regression task, it employs a linear regression head. To adapt CSI-MAE to our classification-based Wi-Fi sensing benchmark, we replace the regression head with a linear classification head while preserving the original linear-probing protocol for a fair comparison.

\begin{figure}[t]
    \centering
    \begin{subfigure}{0.46\textwidth}
        \centering
        \includegraphics[width=\textwidth]{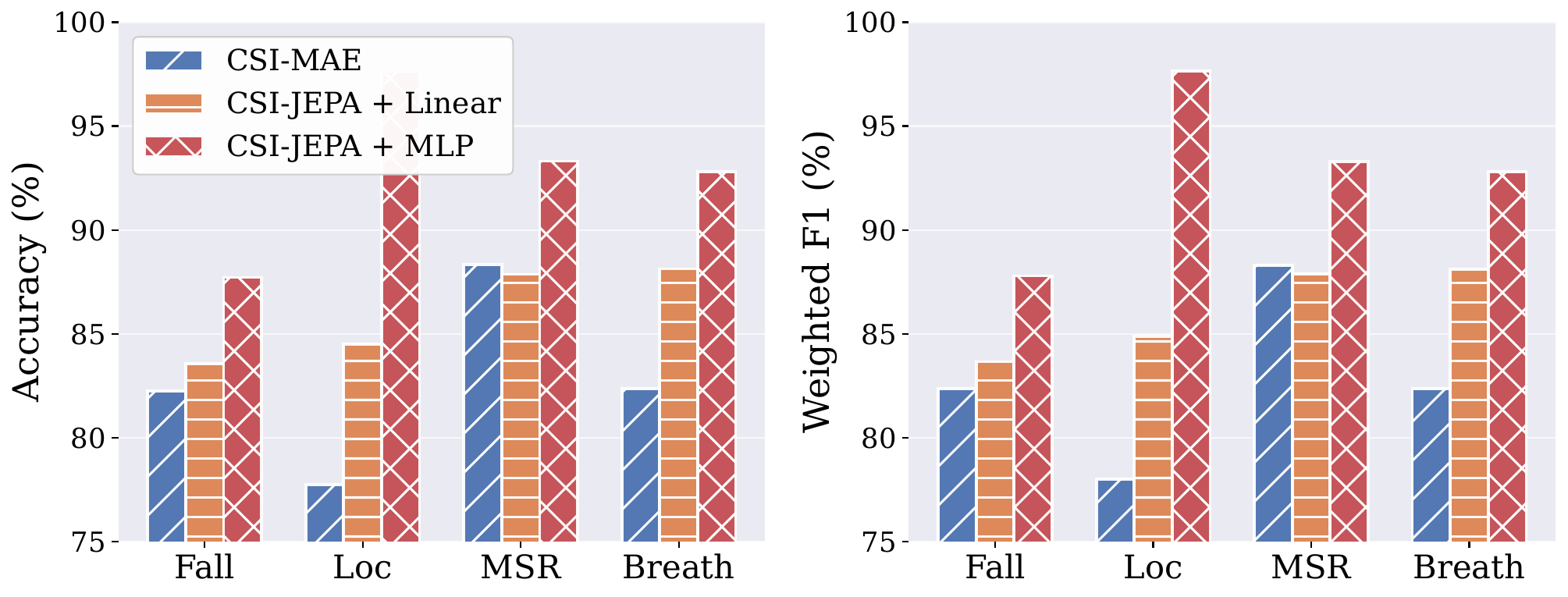}
        \caption{Individually defined sensing tasks.}
    \end{subfigure}
    \hfill
    \begin{subfigure}{0.46\textwidth}
        \centering
        \includegraphics[width=\textwidth]{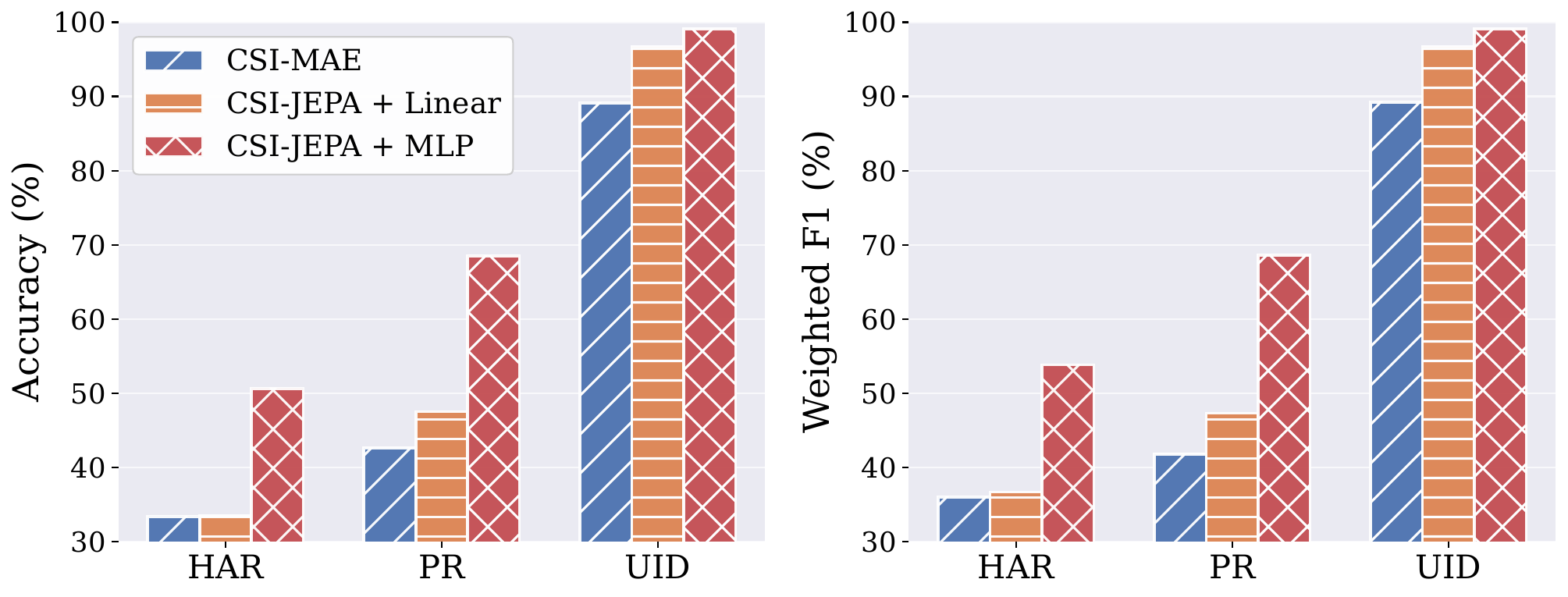}
        \caption{Shared-data sensing tasks}
    \end{subfigure}
    \vspace{-0.5em}
    \caption{Comparison with CSI-MAE under the 1000-label setting. CSI-MAE uses masked CSI reconstruction pretraining, while CSI-JEPA predicts latent target representations.}
    % \vspace{-1em}
    \label{fig:mae_comparison}
\end{figure}

Fig.~\ref{fig:mae_comparison} compares CSI-JEPA with the CSI-MAE-style baseline on individually defined tasks and shared-data tasks. 
CSI-JEPA with a linear probe outperforms CSI-MAE on most tasks, with especially clear gains on Breathing Detection, Localization, Proximity Recognition, and User Identification. 
This indicates that latent-space predictive learning provides more linearly accessible representations than raw masked reconstruction for many CSI-based sensing tasks. 
CSI-MAE slightly outperforms CSI-JEPA Linear on Motion Source Recognition, suggesting that reconstruction-based pretraining can still learn useful channel structure for some cases. 
However, CSI-JEPA with the MLP adapter achieves the best overall performance across all seven sensing tasks, showing that a lightweight nonlinear adapter can further extract task-relevant information from the frozen predictive representation.

\begin{table*}[t]
\centering
\caption{CSI-JEPA backbone ablation across seven tasks. 
\normalfont All backbone variants use the same random temporal-spectral masking strategy. Results are reported as Acc/F1 (\%).}
\label{tab:backbone_mask_ablation}
\small
\begin{tabular}{lcccccccccccccc}
\toprule
\multirow{2}{*}{Backbone}
& \multicolumn{2}{c}{\textbf{Fall}} 
& \multicolumn{2}{c}{\textbf{Loc}} 
& \multicolumn{2}{c}{\textbf{MSR}} 
& \multicolumn{2}{c}{\textbf{Breath}}
& \multicolumn{2}{c}{\textbf{HAR}}
& \multicolumn{2}{c}{\textbf{PR}}
& \multicolumn{2}{c}{\textbf{UID}} \\
\cmidrule(lr){2-3}
\cmidrule(lr){4-5}
\cmidrule(lr){6-7}
\cmidrule(lr){8-9}
\cmidrule(lr){10-11}
\cmidrule(lr){12-13}
\cmidrule(lr){14-15}
& Acc & F1
& Acc & F1
& Acc & F1
& Acc & F1
& Acc & F1
& Acc & F1
& Acc & F1 \\
\midrule
CNN
& 87.72 & 87.69
& 99.64 & 99.63
& 96.42 & 96.42
& 94.39 & 94.39
& 54.09 & 55.38
& 60.88 & 60.29
& 90.45 & 90.46 \\

CNN+ViT
& 85.58 & 85.53
& 98.72 & 98.71
& 96.71 & 96.71
& 94.02 & 94.02
& \textbf{77.18} & \textbf{77.82}
& \textbf{82.99} & \textbf{83.05}
& 99.02 & 99.02 \\

\textbf{ViT} (Ours)
& \textbf{91.27} & \textbf{91.30}
& \textbf{100.00} & \textbf{100.00}
& \textbf{98.41} & \textbf{98.41}
& \textbf{96.87} & \textbf{96.87}
& 76.35 & 77.43
& 81.71 & 81.50
& \textbf{99.88} & \textbf{99.88} \\
\bottomrule
\end{tabular}
\end{table*}

\subsection{Ablation Study on Masking and Backbone}

Next, we investigate how the proposed channel variation-aware masking method and the encoder backbone affect representation quality and downstream sensing performance.
Table~\ref{tab:mask_comparison} first compares four masking strategies under the same ViT backbone and MLP probing protocol. 
The proposed channel-aware masking achieves the best performance on Fall, MSR, HAR, and PR tasks (from Table 1).
Compared with random temporal--spectral masking, channel-aware masking achieves clear improvements on the three tasks where it obtains a $>1$ pp margin, with relative gains ranging from 1.5\% to 6.7\%.
These results suggest that selecting target regions according to local channel variations can provide a more informative predictive task than uniformly sampling target blocks, particularly on tasks where informative CSI patterns exhibit stronger temporal and subcarrier variations.

On Localization and UID tasks, temporal-spectral masking already reaches near-saturated performance, leaving limited room for further improvement. 
This is consistent with the original CSI-Bench results in~\cite{zhu2025csi}, where several supervised models also achieve near-perfect or perfect performance on Room-Level Localization under the official split. 
On Breathing Detection, channel variation-aware masking also remains close to the best alternative, with only a 0.25 pp accuracy gap. These results suggest a task-dependent pattern: channel variation-aware masking provides its clearest gains on sensing tasks with stronger \textit{motion-induced} temporal and subcarrier variations, while remaining close to the best-performing alternative on tasks with relatively weaker or more stationary channel dynamics.

\begin{figure}[t]
    \centerline{\includegraphics[scale=0.35]{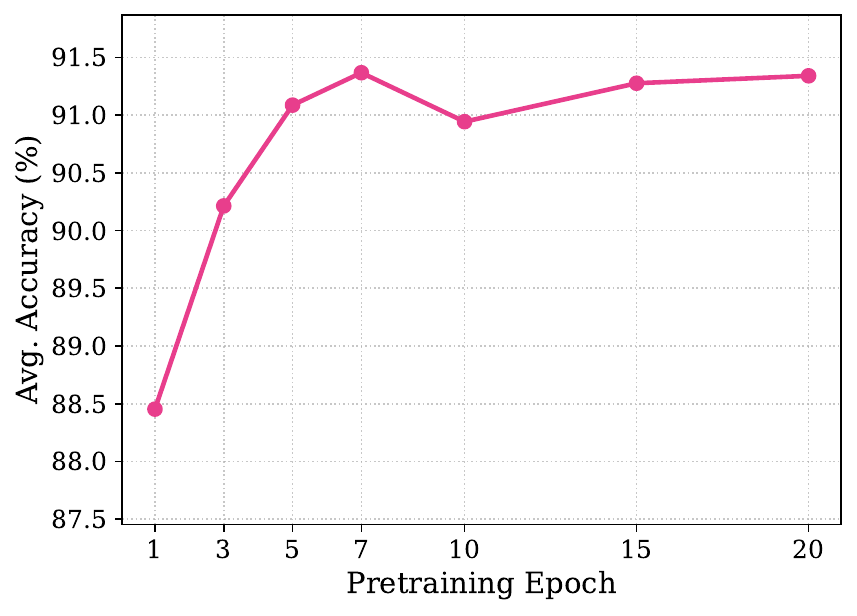}}
    % \vspace{-1em}
    \caption{Effect of pretraining epochs on downstream adaptation. Results are averaged over Fall, MSR, and Breath tasks.}
    \label{fig:epoch}
    \vspace{-1em}
\end{figure}

Besides, Table~\ref{tab:backbone_mask_ablation} compares different encoder backbones under the same random temporal-spectral masking protocol. 
It is observed that the ViT backbone achieves the best performance on most tasks, suggesting that direct self-attention over temporal-subcarrier patch tokens is more effective than CNN-based feature extraction for the pretraining stage. 
Although the CNN-only model uses multiple convolutional stages and produces the same $20\times29$ token grid with 256-dimensional tokens, its convolutional inductive bias mainly emphasizes local patterns before global pooling and projection. 
This can be limiting for CSI-based sensing, where useful cues may involve long-range dependencies across time and subcarriers, such as motion-induced temporal evolution and frequency-selective multipath structures.

Interestingly, the CNN+ViT hybrid model improves over the CNN-only backbone on several tasks, but it still does not outperform the pure ViT backbone. 
This suggests that adding a shallow convolutional front-end does not necessarily improve JEPA-based CSI representation learning, and may distort fine-grained temporal-subcarrier information before the Transformer encoder models global dependencies. 
In contrast, the ViT backbone directly operates on temporal-subcarrier patch tokens and uses self-attention to model global relationships across the entire CSI window. 
Together with the masking ablation in Table~\ref{tab:mask_comparison}, these results support the two main design choices of CSI-JEPA: 1) adopting a ViT-style encoder to capture global temporal–subcarrier dependencies, and 2) employing channel variation–aware masking to focus predictive learning on sensing-informative channel dynamics.

\subsection{Sensitivity and Computational Cost}

\textbf{Effect of pretraining epochs.}
We further evaluate CSI-JEPA checkpoints pretrained for different numbers of epochs using the MLP adapter with 1,000 labeled samples. 
Fig.~\ref{fig:epoch} shows that downstream performance improves rapidly during the first few epochs and becomes stable after about 5--7 epochs. 
Increasing pretraining from 7 to 20 epochs only leads to marginal changes, suggesting that CSI-JEPA can learn useful representations with moderate self-supervised pretraining under the current dataset scale. 

\textbf{Computational Cost.}
Table~\ref{tab:compute_cost} reports the computational cost on Fall Detection with 1,000 labeled samples, measured on a network server with an NVIDIA RTX A6000 GPU. 
Training time is measured for downstream adapter training, and inference latency is measured per CSI window with batch size 1.
Compared with the supervised Transformer, CSI-JEPA MLP updates only 0.067M task-specific adapter parameters, which is about 50$\times$ fewer trainable parameters. 
Although the frozen encoder introduces an additional forward-pass cost during adapter training and inference, the end-to-end latency remains at the millisecond level.
These results indicate that CSI-JEPA does not trade label efficiency for prohibitive computational overhead, and remains practical for online sensing.

\begin{table}[t]
\centering
\caption{Computational cost on the Fall Detection task.} 
\label{tab:compute_cost}
\small
\begin{tabular}{lccc}
\toprule
Method & Params & Training Time & Latency \\
\midrule
Sup. Transformer & 3.351M & 52.47s & 0.522ms \\
CSI-JEPA Linear & 0.001M & 58.54s & 1.287ms \\
CSI-JEPA MLP & 0.067M & 59.41s & 1.297ms \\
\bottomrule
\end{tabular}
% \vspace{-1em}
\end{table}

\section{Conclusion}
This paper presented CSI-JEPA as a step toward reusable foundation representations for Wi-Fi sensing with minimal supervision. Instead of learning a separate supervised model for each sensing task, CSI-JEPA uses masked latent prediction to pretrain a frozen CSI encoder from unlabeled temporal-spectral windows and adapts it to downstream tasks through lightweight heads. The evaluation on seven real-world CSI-Bench tasks shows that this design improves low-label adaptation over state-of-the-art baselines, while achieving matched-budget label savings of up to 95.0\% and maintaining ms-level latency. 
Furthermore, we found that the frozen encoder also transfers to previously unseen sensing tasks on the new dataset under different hardware, environments, and CSI organization, improving cross-dataset adaptation by up to 22.92 pp over the corresponding raw-CSI baseline. These results demonstrate the potential of predictive self-supervised learning for reusable and transferable Wi-Fi sensing representations.

\bibliographystyle{ACM-Reference-Format}
\bibliography{reference}

\end{document}